\documentclass[sigconf]{acmart}

\usepackage{booktabs}   
\usepackage{algorithm}  
\usepackage{algorithmic}
\usepackage{graphicx}   
\usepackage{subcaption} 
\usepackage{multirow}   
\usepackage{enumitem}   
\usepackage{pifont}     
\usepackage{placeins}
\usepackage{tikz}
\usetikzlibrary{arrows.meta, positioning}
\usepackage{pgfplots}
\pgfplotsset{compat=1.18}

\setcopyright{none}
\renewcommand\footnotetextcopyrightpermission[1]{}

\acmConference[KDD '26]{32nd ACM SIGKDD Conference on Knowledge Discovery and Data Mining}{August 9--13, 2026}{Jeju, South Korea}

\title{Reviewing the Reviewer: Graph-Enhanced LLMs for E-commerce Appeal Adjudication}
\author{Yuchen Du}
\email{du313@purdue.edu}
\affiliation{
  \institution{Purdue University}
  \city{West Lafayette}
  \state{Indiana}
  \country{USA}
}

\author{Ashley Li}
\email{ashlili@amazon.com}
\affiliation{
  \institution{Amazon}
  \city{Seattle}
  \state{Washington}
  \country{USA}
}

\author{Zixi Huang}
\email{zixihuan@amazon.com}
\affiliation{
  \institution{Amazon}
  \city{Seattle}
  \state{Washington}
  \country{USA}
}

\begin{abstract}

Hierarchical review workflows, where a second-tier reviewer (Checker) corrects first-tier (Maker) decisions, generate valuable correction signals that encode why initial judgments failed.
However, learning from these signals is hindered by \emph{information asymmetry}: corrections often depend on verification actions unavailable to Makers or automated systems.
We address this challenge by introducing explicit \emph{action modeling} as an inferential constraint that grounds reasoning in verifiable operations rather than unconstrained text generation.
We propose the Evidence-Action-Factor-Decision (EAFD) schema, a minimal representation for adjudication reasoning that prevents hallucination through operational grounding and enables learning from correction signals via explicit conflict modeling.
Building on this schema, we develop a conflict-aware graph reasoning framework that: (1) constructs EAFD graphs from historical cases capturing Maker-Checker disagreements, (2) aggregates them into a retrievable knowledge base, and (3) performs top-down deductive reasoning for new cases by projecting validated resolution paths from precedents.
A distinctive capability is the \emph{Request More Information} (RMI) outcome: when evidence is insufficient, the system identifies precisely which verification actions remain unexecuted and generates targeted information requests.
We evaluate the framework in large-scale e-commerce seller appeal adjudication.
While a standard LLM-only baseline achieves only 70.8\% alignment with human experts, incorporating action modeling with RMI improves alignment to 87.5\%.
Augmenting this with the retrieval-based knowledge graph yields the best offline performance of 95.8\%.
Following online deployment, the framework maintains robust performance, achieving a 96.3\% alignment rate in production, demonstrating its real-world effectiveness.

\end{abstract}



\definecolor{approvegreen}{RGB}{220, 252, 231}
\definecolor{rejectred}{RGB}{254, 226, 226}
\definecolor{rmiyellow}{RGB}{254, 249, 195}
\newcommand{\approve}{\colorbox{approvegreen}{\textsc{Approve}}}
\newcommand{\reject}{\colorbox{rejectred}{\textsc{Reject}}}
\newcommand{\rmi}{\colorbox{rmiyellow}{\textsc{RMI}}}

\AtBeginDocument{
  \abovedisplayskip=4pt plus 2pt minus 2pt
  \belowdisplayskip=4pt plus 2pt minus 2pt
  \abovedisplayshortskip=0pt plus 2pt
  \belowdisplayshortskip=2pt plus 2pt minus 2pt
}

\begin{document}

\maketitle

\section{Introduction}
\label{sec:intro}


In high-stakes decision domains, particularly e-commerce dispute resolution, platforms face a dual challenge: maintaining strict compliance while preventing wrongful restrictions on legitimate sellers \cite{lin2025fraudulent, chen2020review}.
To balance these objectives, organizations commonly deploy hierarchical \emph{Maker-Checker} review workflows \cite{shi2025antakso, kondo2025capturing}.
In this setting, a first-tier reviewer (Maker) screens seller appeals and issues initial rejections; a second-tier reviewer (Checker) then re-examines these rejected cases and may overturn the decision based on deeper investigation.
These \emph{correction events}, in which an initial judgment is overturned upon deeper inspection, represent a uniquely valuable but underexploited learning signal \cite{dsouza2025sources}.
Unlike standard labeled data that only encodes \emph{what} decision was reached, corrections encode \emph{why an initial judgment failed}, revealing specific reasoning gaps, evidence misinterpretations, and verification shortcuts.
In principle, capturing this corrective logic could allow automated systems to reduce false positives and prevent unnecessary escalations \cite{hatalis2025review}.

However, exploiting these correction signals is hindered by the profound \emph{information asymmetry} inherent in hierarchical reviews.
Checkers often overturn Maker decisions not because they reason better over the same information, but because they perform additional verification actions (e.g., consulting external supplier databases, validating invoice authenticity, or cross-referencing transaction logs) that the Maker omitted \cite{kondo2025capturing}.
This asymmetry is exacerbated by the heterogeneous nature of e-commerce evidence: a single seller appeal often involves multi-turn interactions with loosely structured materials including free-form text explanations, scanned invoices, chat logs, and images \cite{shah2025towards}.
Reviewers must extract relevant signals from noisy data without fully standardized operating procedures, leading to decisions that depend heavily on individual experience \cite{chen2020review}.
Consequently, what appears as a ``reasoning error'' by the Maker is often an ``information gap.''
Standard machine learning approaches, including Large Language Models (LLMs), struggle in this setting: when asked to predict the Checker's decision directly from the Maker's partial evidence, they face an impossible inference task \cite{hatalis2025review, wei2022chain}.
Without modeling the missing verification steps, LLMs tend to \emph{hallucinate} plausible-sounding rationales, such as asserting ``the supplier documentation appears authentic'' without any verification mechanism, or claiming ``the transaction pattern is consistent with legitimate business'' without defining what consistency check was performed \cite{chen2024sac, jin2024graph}.
Such outputs may be linguistically coherent but are operationally meaningless in adjudication contexts where every claim must be traceable to specific evidence.

Our key insight is that bridging this gap requires an \emph{action-centric} approach: modeling not just what reviewers concluded, but what they \emph{did} to reach those conclusions.
We observe that expert adjudication is not a direct jump from evidence to decision, but a sequence of \emph{verification actions} (e.g., checking supplier registries, validating invoice authenticity, cross-referencing transaction records) that transform static evidence into validated factors \cite{wiratunga2024cbr}.
By explicitly modeling this action layer, we impose an \emph{inferential constraint}: the system cannot simply predict an outcome; it must specify \emph{which verification action} was performed on \emph{which evidence} to yield \emph{what outcome} \cite{aamodt1994case, kolodner1992introduction}.
This structural grounding transforms the learning problem from unconstrained text generation into structured reasoning over verifiable operations \cite{jin2024graph}.
The action layer enables the system to distinguish between cases that need better reasoning (Factor Conflict) and cases that simply need more data (Action Gap).
This enables a \emph{Request More Information} (RMI) capability: instead of forcing a low-confidence guess, the system can identify precisely which verification actions remain unexecuted and recommend them to human reviewers. For instance, it may output: ``Unable to verify supplier authenticity; please provide the original purchase invoice from the claimed supplier.''
This ``know what you don't know'' capability transforms the system from a black-box predictor into an actionable decision support tool \cite{guo2024ds}.

To operationalize this insight, we propose a conflict-aware reasoning framework centered on the Evidence-Action-Factor-Decision (EAFD) schema. This representation explicitly separates raw materials (Evidence), verification procedures (Actions), abstract criteria (Factors), and final outcomes (Decisions), enabling the system to learn directly from Maker-Checker disagreements through explicit conflict edges \cite{patel2025graph, dsouza2025sources}.
We instantiate this framework in e-commerce seller appeal adjudication, focusing specifically on \textbf{Maker-rejected appeals} where the Checker's role is to identify whether the initial rejection was warranted.
Our goal is to produce decisions aligned with the Checker's judgment. However, due to inherent information asymmetry, we adopt a \emph{conservative strategy}: when evidence is insufficient, the system outputs a Request More Information (RMI) decision along with recommended verification actions and evidence to review, rather than forcing a potentially incorrect judgment.

This paper makes the following contributions:
\begin{itemize}[leftmargin=*, nosep]
    \item \textbf{Problem Formulation.} We identify information asymmetry as the primary bottleneck in learning from hierarchical review corrections and formalize the need for action-level modeling to mitigate hallucination.
    \item \textbf{EAFD Schema.} We introduce a novel graph schema that bridges unstructured evidence and adjudication policies, serving as a structured medium to capture correction signals while preventing hallucination through operational grounding.
    \item \textbf{Conflict-Aware Reasoning Framework.} We develop a graph-based system that constructs EAFD graphs from historical cases, aggregates them into a retrievable knowledge base, and performs top-down deductive reasoning (Factor$\to$Action$\to$Evidence) for new cases by projecting validated resolution paths from precedents.
    \item \textbf{Industrial Deployment.} We deploy the system in a large-scale e-commerce platform. Offline evaluation shows that our framework achieves \textbf{95.8\% accuracy}, significantly outperforming the standard LLM baseline of 70.8\%. Following online deployment, the system demonstrates robust real-world effectiveness with a \textbf{96.3\% cumulative alignment rate} and maintains \textbf{100\% precision on approvals}, ensuring no wrongful reinstatements occur in high-stakes risk management scenarios.
\end{itemize}

The remainder of this paper is organized as follows. Section~\ref{sec:lr} reviews related work. Section~\ref{sec:problem} formalizes the problem setting. Section~\ref{sec:method} presents the EAFD framework. Section~\ref{sec:exp} reports experimental results. Section~\ref{sec:conc} presents lessons learned, and Section~\ref{sec:conclusion} concludes.

\begin{figure*}[t]
    \centering
    \includegraphics[width=0.95\textwidth]{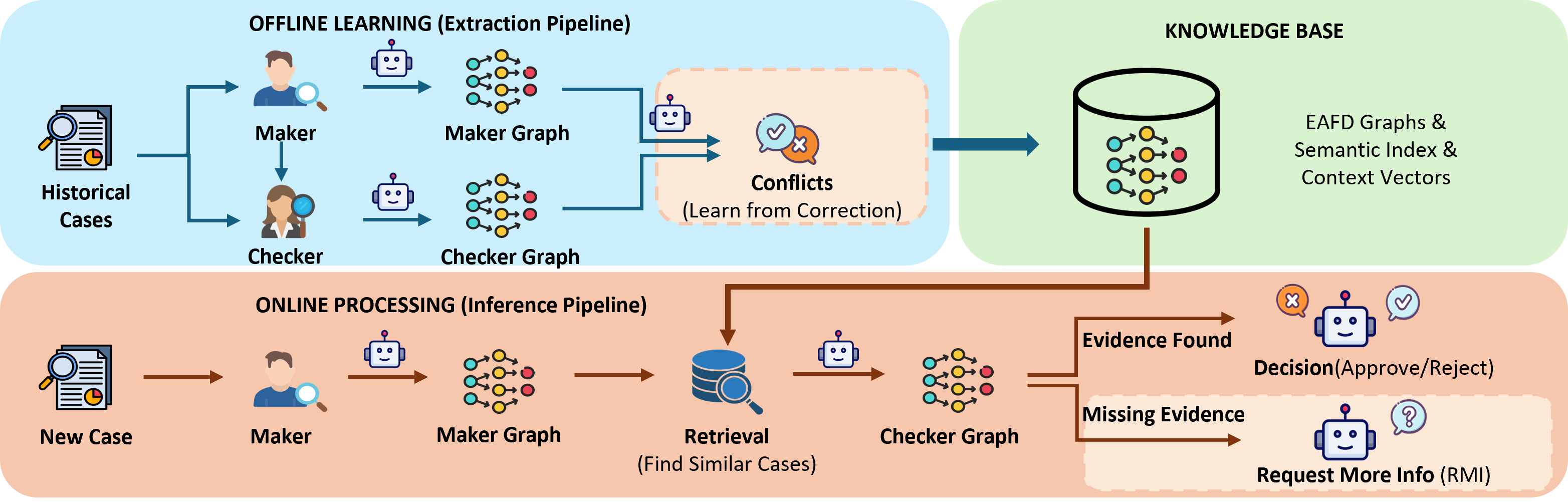}
    \caption{System overview. \textbf{Offline:} Extract EAFD graphs from historical cases and build a knowledge base. \textbf{Online:} Construct Maker graph, retrieve precedents, derive Checker graph via FAE deduction, and output adjudication decision.}
    \label{fig:system_overview}
\end{figure*}

\section{Related Work}
\label{sec:lr}

\paragraph{Case-Based Reasoning.}
Case-Based Reasoning (CBR) traditionally solves new problems by retrieving and adapting solutions from similar past cases, offering high reliability but limited flexibility when handling unstructured or heterogeneous data \cite{aamodt1994case, das2020simple, ning2024urbankgent, xu2023knowledge}. 
Recent work has augmented the CBR cycle with Large Language Models (LLMs), substantially improving adaptation and reasoning capabilities in domains such as legal question answering, software testing, and safety-critical decision support \cite{gan2025case, guo2025optimizing, hatalis2025review, wiratunga2024cbr, shen2025law, guo2024ds}. 
However, applying LLM-augmented CBR to high-stakes industrial adjudication exposes a critical veracity gap, as generated content may diverge from factual evidence or lack sufficient domain grounding. 
Without explicit structural constraints, LLMs are prone to hallucinating facts or misinterpreting domain rules, leading to unreliable decisions \cite{chen2024sac, kondo2025capturing, hatalis2025review, guo2024ds}. 
To address this limitation, our approach introduces explicit Action modeling into the reasoning and adaptation process, grounding LLM reasoning in verifiable operational behaviors rather than unstructured text and thereby enforcing adherence to validated procedural logic.

\paragraph{Structured Reasoning.}
Integrating Knowledge Graphs (KGs) with LLMs has emerged as a dominant paradigm for improving reasoning robustness \cite{chen2020review, jin2024graph, patel2025graph, he2024give, jiang2025kg, tan2024struct}. 
Methods such as Graph Chain-of-Thought (Graph-CoT) exploit graph structures to decompose complex queries into intermediate reasoning steps \cite{jin2024graph}, and recent work has extended these ideas to incomplete KG question answering and automatic graph construction from unstructured text \cite{chen2024sac, shan2025automatic, xu2024generate, zhao2025agree}. 
Nevertheless, most existing Graph RAG approaches focus on modeling static factual relationships, such as entity attributes or taxonomic links \cite{chen2020review, patel2025graph, he2024give}. 
In contrast, dispute adjudication is inherently procedural, requiring explicit modeling of how evidence is verified and aggregated over time to support a decision \cite{kondo2025capturing, shen2025law}. 
Our work addresses this gap through the EAFD schema, which represents adjudication as navigable reasoning paths (Evidence $\to$ Action $\to$ Factor), enabling the system to trace coherent procedural flows from raw evidence to final judgments. Unlike preference-based or outcome-level supervision, our framework treats reviewer corrections as structured reasoning transformations, explicitly modeling how Checker actions and factors invalidate Maker judgments.

\paragraph{Learning from Human Review Corrections.}
Human-in-the-loop learning and Reinforcement Learning from Human Feedback (RLHF) are widely used to align model behavior with human preferences \cite{xu2025rlthf}, but they typically optimize toward a single final decision or scalar feedback signal. 
In hierarchical industrial workflows, however, the effective ground truth often emerges through a Maker--Checker process, in which Checkers correct Maker judgments rather than merely expressing preference over outcomes. 
Despite its prevalence in practice, this correction signal is rarely modeled explicitly in prior work: overturned cases are commonly treated as noise or reduced to binary labels, discarding the meta-reasoning embedded in the disagreement itself \cite{dsouza2025sources}. 

In contrast to RLHF, which updates model parameters through algorithmic training, our work leverages human corrections at inference time as structured reasoning signals without modifying the underlying LLM. 
Specifically, reviewer disagreement is treated as a form of supervision over \emph{reasoning structure} rather than over final decisions alone. 
By incorporating an explicit conflict detection mechanism, the system transforms historical Maker--Checker conflicts into structured learning signals, enabling meta-reasoning that captures not only what decision was reached, but why an initial judgment failed. 
This explicit modeling of overruling logic addresses a form of procedural knowledge that remains challenging for LLMs to acquire implicitly \cite{zhang2025llms}.

\section{Problem Formulation}
\label{sec:problem}

We formalize the problem of learning from hierarchical review corrections as a structured reasoning task.

Let $\mathcal{C} = \{c_1, c_2, \ldots, c_N\}$ denote a dataset of historical appeal cases. Each case $c_i$ is a tuple $(\mathcal{E}_i, d_i^{(M)}, d_i^{(C)})$, where:
\begin{itemize}[leftmargin=*, nosep]
    \item $\mathcal{E}_i$ represents the \textbf{unstructured case context}, including seller metadata, violation details, and multi-modal evidence (e.g., invoices, chat logs, product images).
    \item $d_i^{(M)} \in \{\reject\}$ is the \textbf{Maker's decision record}, comprising both the decision outcome and the accompanying analysis text (evidence examination, verification steps taken, and rationale for rejection). Consistent with our scope (Section~\ref{sec:intro}), we focus on cases initially rejected by the Maker.
    \item $d_i^{(C)} \in \{\approve, \reject\}$ is the \textbf{Checker's decision record}, similarly containing the final decision along with detailed reasoning that documents what additional verifications were performed and why the Maker's judgment was upheld or overturned. This serves as the ground truth.
\end{itemize}

\noindent\textbf{Task Definition.}
Given a new query case $c_q$ with evidence $\mathcal{E}_q$ and the Maker's rejection $d_q^{(M)}$, the system aims to produce a decision $\hat{d}_q^{(C)} \in \{\approve, \reject, \rmi\}$ aligned with the Checker's judgment, along with a reasoning path grounded in verification actions.

The core challenge is \emph{information asymmetry}: the transition from $d^{(M)}$ to $d^{(C)}$ often depends on verification actions $\mathcal{A}$ not explicitly present in the raw evidence $\mathcal{E}$.
To address this, we adopt a conservative strategy. The objective is to learn a mapping:
\begin{equation}
f: (\mathcal{E}_q, d_q^{(M)}) \to (\hat{d}_q^{(C)}, \mathcal{A}_{rec}),
\end{equation}
where $\mathcal{A}_{rec}$ denotes recommended verification actions. When evidence is insufficient to support a confident \approve{} or \reject{}, the system outputs $\hat{d}_q^{(C)} = \rmi$ with $\mathcal{A}_{rec}$ specifying which actions remain unexecuted and what evidence is needed.

\section{Methodology}
\label{sec:method}

We propose a Conflict-Aware Graph Reasoning framework that automates adjudication by explicitly modeling the divergence between Maker rejections and Checker verifications.
As illustrated in Figure~\ref{fig:system_overview}, the methodology comprises three components.
First, we define the \textbf{EAFD Graph Schema} (Section~\ref{sec:eafd_schema}), a structured representation capturing how evidence, actions, and factors interact to form decisions.
Second, we describe \textbf{Offline Knowledge Base Construction} (Section~\ref{sec:kb_construction}), which extracts EAFD graphs from historical cases and aggregates them into a semantically indexed knowledge base.
Finally, we present \textbf{Online Conflict-Aware Reasoning} (Section~\ref{sec:online_reasoning}), a deductive inference engine that retrieves precedents to construct verifiable resolution paths for new appeals.

%

\subsection{EAFD Graph Schema}
\label{sec:eafd_schema}

We represent each appeal case $c$ as a directed, typed graph $\mathcal{G}_c = (\mathcal{V}_c, \mathcal{R}_c)$ with an explicit two-tier review structure $
\mathcal{G}_c = \mathcal{G}_c^{(M)} \cup \mathcal{G}_c^{(C)} $,
where $\mathcal{G}_c^{(M)}$ captures the Maker's initial review and $\mathcal{G}_c^{(C)}$ captures the Checker's subsequent review. The two subgraphs form parallel reasoning traces connected by cross-graph \emph{conflict edges} that encode Maker--Checker disagreement. Following Section~\ref{sec:problem}, $\mathcal{G}_c^{(M)}$ is constructed from $(\mathcal{E}_c, d_c^{(M)})$ and $\mathcal{G}_c^{(C)}$ from $(\mathcal{E}_c, d_c^{(C)})$.

\begin{figure}[t]
    \centering
    \includegraphics[width=1.05\columnwidth]{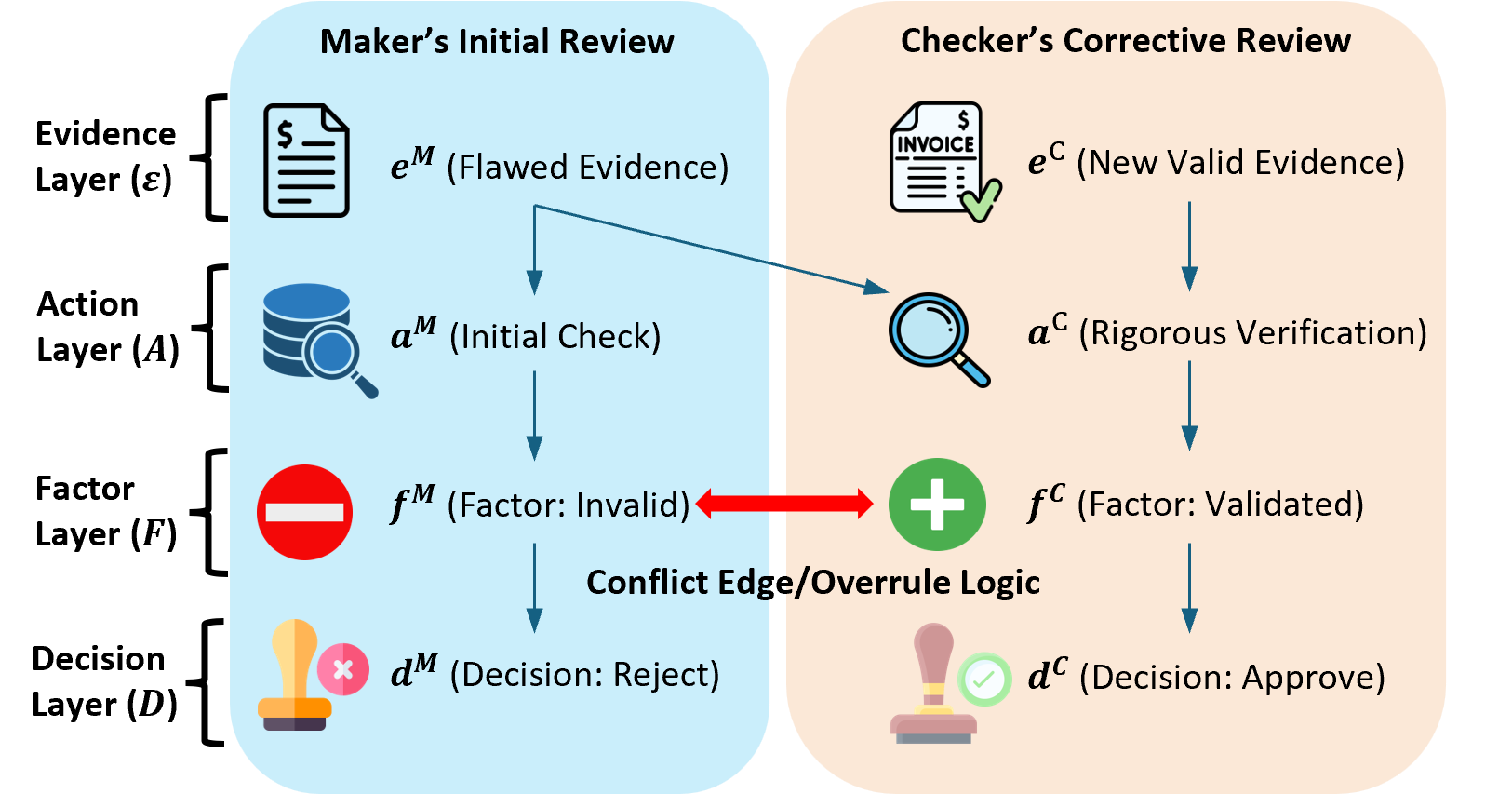}
    \vspace{-8pt}
    \caption{The EAFD schema with four-layer hierarchy. Left: Maker's initial review leads to rejection. Right: Checker's corrective review with additional verification overturns the decision. The conflict edge (red) captures Maker-Checker disagreement as a learning signal.}
    \label{fig:eafd_schema}
\end{figure}

\paragraph{Node Types (structure).}
The node set $\mathcal{V}_c = \mathcal{E}_c \cup \mathcal{A}_c \cup \mathcal{F}_c \cup \mathcal{D}_c $
comprises four functional layers: \textbf{Evidence} nodes $\mathcal{E}_c$ (atomic facts with source references), \textbf{Action} nodes $\mathcal{A}_c$ (verification operations documented in decision records), \textbf{Factor} nodes $\mathcal{F}_c$ (abstract decision criteria), and \textbf{Decision} nodes $\mathcal{D}_c$ (Maker/Checker outcomes). Figure~\ref{fig:eafd_schema} illustrates an instantiated example, and Appendix~\ref{app:eafd_details} provides formal specifications.

\paragraph{Relation Types (structure).}
Edges in $\mathcal{R}_c$ encode procedural dependencies:
Evidence--Action ($\mathcal{R}_{EA}$, many-to-many), Action--Factor ($\mathcal{R}_{AF}$, one-to-one), Factor--Decision ($\mathcal{R}_{FD}$), and cross-graph Factor--Factor ($\mathcal{R}_{FF} \subseteq \mathcal{F}_c^{(M)} \times \mathcal{F}_c^{(C)}$). The core design choice is that $\mathcal{R}_{FF}$ explicitly models how Checker-level reasoning evaluates, overrides, or invalidates Maker-level factors, making disagreement a first-class supervision signal rather than an unstructured artifact.

\vspace{-8pt}

\paragraph{Action grounding (semantics and constraint).}
A distinctive feature of EAFD is that \textbf{Action nodes act as inferential constraints}: they ground factors in \emph{verifiable operations} rather than unconstrained text generation, mitigating hallucinations in high-stakes adjudication. Actions are \emph{not} sampled from a predefined taxonomy; instead, they are \emph{extracted} from reviewer decision records $d^{(M)}$ and $d^{(C)}$, where verification behaviors are explicitly documented. Each action corresponds to a \emph{goal-oriented verification behavior}. For example, verifying supplier legitimacy may involve multiple operational methods (e.g., registry lookup, document cross-reference, third-party contact), but these constitute a single Action because they serve the same verification goal.

This extraction-based design addresses a key failure mode of LLM-only adjudication: if actions were freely generated, they could themselves be hallucinated. By constraining actions to recurring patterns recoverable from historical decision logs and requiring each action to be grounded in specific evidence items via $\mathcal{R}_{EA}$, EAFD ensures that actions remain operationally meaningful rather than merely linguistically plausible. During offline learning, actions sharing the same verification goal and factor association can be merged, yielding a compact set of recurring operation types.

\paragraph{Factor abstraction (interpretation layer).}
Factor nodes represent \emph{interpretable decision criteria} derived from action outcomes, bridging concrete verification results and final adjudication. While an Action answers \emph{what was verified and how}, a Factor answers \emph{what the verification result implies for the decision}. For instance, the action ``cross-reference invoice with supplier registry'' may yield ``Supplier Relationship Confirmed'' (supporting approval) or ``Supplier Authenticity Unverified'' (supporting rejection), depending on the verification outcome.

Each factor $f \in \mathcal{F}_c$ is associated with a semantic outcome attribute $o(f) \in \{\text{Support}, \text{Contradict}\}$, linking the verification result to decision $d$. EAFD models Maker--Checker disagreement at the \textbf{Factor level} rather than at the Evidence or Action level. Two reviewers may inspect the same evidence and even perform similar verification actions, yet form different factors due to interpretive differences: one may interpret an ambiguous registry match as ``insufficient proof'' while another considers it ``reasonable confirmation.'' By localizing conflict to the factor abstraction layer, the framework preserves the interpretive nature of adjudication while maintaining traceability down to the underlying actions and evidence.

Factors are intentionally domain- and policy-specific rather than drawn from a universal ontology. Their semantics may evolve with compliance rules, risk tolerance, and operational objectives, allowing the same verification actions to support different decision logics under different factor definitions (see Section~\ref{sec:conc}).

\begin{figure*}[t]
    \centering
    \includegraphics[width=0.95\textwidth]{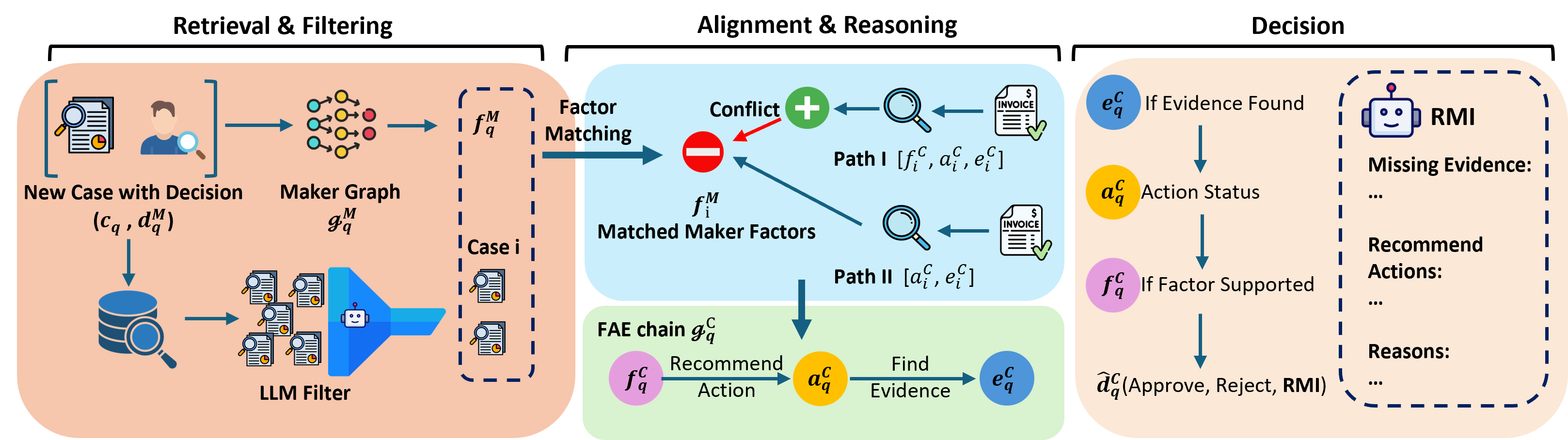}
    \caption{Online reasoning pipeline. The system constructs the Maker graph, retrieves similar cases, aligns factors to historical paths, and applies FAE deduction ($f_q^{(C)} \to a_q^{(C)} \to e_q^{(C)}$). Final decision $\hat{d}_q^{(C)}$ is based on action status; RMI is triggered when critical evidence is missing.}
    \label{fig:online_reasoning}
\end{figure*}

\subsection{Offline Knowledge Base Construction}
\label{sec:kb_construction}





The offline pipeline populates the domain-specific Knowledge Base (KB) by distilling adjudication logic from historical review records. Rather than constructing a static factual knowledge graph, the pipeline captures procedural adjudication logic embedded in Maker and Checker review traces. This process transforms heterogeneous text records into structured, queryable reasoning subgraphs that preserve the corrective insights of expert reviewers.

\paragraph{Hierarchical Graph Extraction.}For each historical case $c_i$, the system constructs the EAFD graph $\mathcal{G}_i$ through a multi-stage, LLM-driven top-down extraction process. This top-down order mirrors how adjudication reasoning is documented in professional environments: decisions cite factors, factors are justified by actions, and actions reference evidence.
\begin{enumerate}
    \item \emph{Decision \& Factor Identification}: The system identifies the categorical outcomes $d_i^{(M)}$ and $d_i^{(C)}$, and extracts the underlying factors $\mathcal{F}_{c_i}$ categorized by their source logs (Maker vs. Checker).
    \item \emph{Action Grounding}: For each factor $f$, the system identifies the set of verification actions $\mathcal{A}_{i}(f)$ performed to reach that outcome. \textbf{Actions are restricted to those explicitly documented in review records, preventing the introduction of unverifiable operations.}
    \item \emph{Evidence Linking}: Each action is grounded in atomic text snippets $\mathcal{E}_{i}(a) \subseteq \mathcal{E}_i$, ensuring that every reasoning step is traceable to the original source evidence.
\end{enumerate}

\paragraph{Action Refinement and Merging.}To address the linguistic variability in how different reviewers document verification, the system performs \textbf{Action Refinement}. Multiple operations sharing the same functional goal and factor association are merged into canonical Action nodes (e.g., "manual registry check" and "portal lookup" are unified into "Verify Supplier Legitimacy"). This step reduces graph redundancy and ensures that downstream retrieval operates over stable verification primitives rather than case-specific phrasing.

\paragraph{Structural Invariant Validation.}To prevent the propagation of malformed reasoning into the KB, a \textbf{Validation Layer} enforces structural invariants on the generated graphs. Specifically, the layer enforces: (1) \emph{Type Compatibility} (e.g., edges must only connect permitted layers); (2) \emph{Relational Completeness} (e.g., every factor must be grounded in an action); and (3) \emph{Cardinality Invariants} (e.g., the 1-to-1 mapping between Actions and Factors). This validation layer decouples probabilistic semantic extraction from deterministic structural correctness, triggering targeted regeneration for any detected inconsistencies.

\paragraph{Knowledge Modeling and Indexing.}For overturned cases ($d_i^{(M)} \neq d_i^{(C)}$), the system explicitly constructs conflict edges $\mathcal{R}_{FF}$ representing the correction signal that links erroneous Maker factors to validated Checker reasoning paths. This encodes the "why" behind rejections being overturned. To enable efficient retrieval, each graph is indexed by a dense semantic vector $\mathbf{z}_i$ derived via:
\begin{equation}c_i \xrightarrow{\text{LLM}} s_i \xrightarrow{\text{Encoder}} \mathbf{z}_i,
\end{equation} where $s_i$ is a structured summary encapsulating the violation category and core rationale. The final KB is stored as $\mathcal{K} = \{(\mathcal{G}_i, \mathbf{z}_i)\}_{i=1}^N$, supporting efficient similarity-based retrieval of historical resolution paths.

\subsection{Online Conflict-Aware Reasoning}
\label{sec:online_reasoning}

When a new appeal case $c_q$ arrives, the system predicts the Checker's decision $\hat{d}_q^{(C)}$ by resolving conflicts in the Maker's initial judgment. Rather than directly inferring an outcome from raw evidence, the system follows a structured \emph{retrieval--deduction} pipeline that projects historically validated resolution paths onto the current case.

\paragraph{Maker Graph Construction.}
The system first constructs the Maker-level subgraph $\mathcal{G}_q^{(M)}$ from raw case materials and the initial rejection decision $d_q^{(M)}$. This step identifies the set of Maker factors $\mathcal{F}_q^{(M)}$ that explicitly define the grounds for rejection and serve as the core conflict points to be examined during online reasoning.

\paragraph{Retrieval and Factor Alignment.}
To identify precedents that offer actionable guidance, the system performs a coarse-to-fine retrieval process. First, the system generates a structured summary $s_q$ of the case context and encodes it into a query embedding $\mathbf{z}_q$. 
An embedding-based retriever performs coarse-grained candidate selection from the knowledge base:
\begin{equation}
\mathcal{C}_{\text{cand}} =
\operatorname{TopK}_{(\mathcal{G}_i, \mathbf{z}_i) \in \mathcal{K}}
\; \cos(\mathbf{z}_q, \mathbf{z}_i).
\end{equation}
These candidates provide a recall-oriented superset that bounds the subsequent reasoning scope.
An LLM-based refiner then performs semantic filtering over $\mathcal{C}_{\text{cand}}$ by comparing structured summaries, producing a refined set of analogous cases $\mathcal{C}_{\text{sim}}$, which is used for all downstream factor alignment and reasoning.

Retrieving similar cases alone is insufficient: the system must further identify \emph{which} historical reasoning elements are relevant to the current dispute. Accordingly, retrieval operates at two levels: \emph{case-level} similarity for context narrowing, followed by \emph{factor-level} alignment for reasoning transfer. For each Maker factor in $\mathcal{F}_q^{(M)}$, the system aligns it with semantically equivalent historical Maker factors:
\begin{equation}
\begin{split}
\mathcal{F}_{\text{aligned}} = \bigl\{ f_i^{(M)} \in \mathcal{G}_i \mid{} & c_i \in \mathcal{C}_{\text{sim}},\\
& \exists f_q^{(M)} \in \mathcal{F}_q^{(M)} : \operatorname{Match}(f_q^{(M)}, f_i^{(M)}) \bigr\}
\end{split}
\end{equation}
where $\operatorname{Match}(\cdot)$ denotes LLM-based semantic alignment. The resulting set $\mathcal{F}_{\text{aligned}}$ serves as a collection of \emph{historical anchors} from which the system traces validated verification paths.

\paragraph{FAE Deductive Inference.}
Based on these anchors, the system constructs the Checker's reasoning subgraph $\mathcal{G}_q^{(C)}$ as a parallel reasoning path that explicitly addresses conflicts in $\mathcal{G}_q^{(M)}$. As illustrated in Figure~\ref{fig:online_reasoning}, this process is guided by Factor--Factor conflict edges in $\mathcal{R}_{FF}$.

For each aligned historical Maker factor $f_i^{(M)} \in \mathcal{F}_{\text{aligned}}$, the system extracts connected Checker-level reasoning structures from historical graphs, which typically take one of the following forms:
\begin{itemize}[leftmargin=*, nosep]
    \item \textbf{Path I (Verification)}: $e_i \rightarrow a_i^{(C)} \rightarrow f_i^{(M)}$, where the Checker's action directly verifies the Maker's factor.
    \item \textbf{Path II (Extension)}: $e_i \rightarrow a_i^{(C)} \rightarrow f_i^{(C)} \leftrightarrow f_i^{(M)}$, where the Checker introduces a new factor that conflicts with or overrides the Maker's factor.
\end{itemize}

Unlike the bottom-up extraction process used during offline learning, online reasoning follows a top-down \textbf{Factor--Action--Evidence (FAE)} deductive chain to instantiate these structures onto the current case:
\begin{itemize}[leftmargin=*, nosep]
    \item \textbf{(F) Factor Hypothesizing.} Guided by historical resolution paths, the system instantiates candidate Checker factors $f_q^{(C)}$ corresponding to each $f_q^{(M)}$. Factors are selected from those that successfully resolved similar conflicts in precedent cases, and a conflict edge $(f_q^{(M)}, f_q^{(C)}) \in \mathcal{R}_{FF}$ is established to guide subsequent verification.
    \item \textbf{(A) Action Planning.} The system deduces required verification actions $a_q^{(C)}$ by adapting historical actions $a_i^{(C)}$ to the current case context. Rather than copying actions verbatim, semantic adaptation replaces case-specific entities (e.g., ``verify supplier $X$ in registry $Y$'' becomes ``verify supplier $X'$ in registry $Y'$'').
    \item \textbf{(E) Evidence Grounding.} Actions are grounded by searching $\mathcal{E}_q$ for supporting evidence traces. The system distinguishes three outcomes: \emph{complete match} (evidence fully supports the action), \emph{partial match} (evidence is relevant but inconclusive), and \emph{missing} (no relevant evidence found). Only complete matches yield \textsc{Verified} status.
\end{itemize}
Through this FAE sequence, the system constructs a grounded reasoning graph in which every instantiated node is both historically precedented and explicitly supported by the current case data.

\paragraph{Adjudication and RMI.}
The final stage derives a structured adjudication outcome from the instantiated graph $\mathcal{G}_q^{(C)}$. We define the action status function:
\begin{equation}
\mathcal{S}(a^{(C)}) =
\begin{cases}
\textsc{Verified}, & \text{if } \exists\, e \in \mathcal{E}_q \text{ s.t. } (e, a^{(C)}) \in \mathcal{R}_{EA}, \\
\textsc{Missing}, & \text{otherwise}.
\end{cases}
\end{equation}

Action status further determines whether a factor's semantic outcome is \emph{actionable} in decision making. 
A factor $f_q^{(C)}$ whose supporting actions are all marked as \textsc{Verified} contributes its outcome $o(f_q^{(C)}) \in \{\text{Support}, \text{Contradict}\}$ to adjudication. 
Conversely, if any critical action required to support $f_q^{(C)}$ is marked as \textsc{Missing}, the factor is treated as \emph{unresolved} and does not exert decisive influence on the final decision.

Finally, the system performs a holistic assessment over the annotated graph \emph{under the structural constraints imposed by verified and missing actions}, producing a decision $\hat{d}_q^{(C)} \in \{\approve, \reject, \\ \rmi\}$.
When critical actions are marked as \textsc{Missing}, the system outputs an RMI decision together with a recommendation set:
\begin{equation}
\mathcal{A}_{\text{rec}} = \bigl\{ a^{(C)} \in \mathcal{G}_q^{(C)} \mid{} \mathcal{S}(a^{(C)}) = \textsc{Missing} \land\; \operatorname{Critical}(a^{(C)}) \bigr\}.
\end{equation}
Each recommended action is translated into a targeted information request (e.g., ``Please provide the supplier registration certificate'').

Not all missing actions carry equal weight. The system distinguishes \emph{critical actions}, whose absence blocks any approval, from \emph{supporting actions}, whose absence weakens but does not preclude approval. A case with all critical actions verified but some supporting actions missing may still receive \approve{}, whereas the absence of any critical action triggers \rmi{} with explicit recommendations.

Unlike confidence-threshold approaches that abstain when model certainty is low, the proposed RMI mechanism is \emph{structurally driven}. Historical precedents indicate that resolving a factor $f$ requires executing a specific action $a$; if no evidence in $\mathcal{E}_q$ can ground $a$ (i.e., $\mathcal{S}(a)=\textsc{Missing}$), the system identifies an information gap rather than a reasoning failure. This structural grounding enables precise, actionable information requests instead of generic uncertainty statements.

\section{Experiments}
\FloatBarrier
\label{sec:exp}

\subsection{Experimental Setup}
\subsubsection{Datasets}

We evaluate our framework on a high-stakes seller appeal dataset collected from a large-scale e-commerce risk management platform. Unlike standard classification benchmarks, each case corresponds to a complex, multi-turn adjudication process involving heterogeneous and multi-modal evidence (e.g., invoices, chat logs, and product images) and hierarchical expert review. The dataset captures the full lifecycle of Maker–Checker interactions, where initial rejections by first-tier reviewers (Makers) are frequently revisited and overturned by second-tier experts (Checkers) through deeper verification actions, providing rich and explicit correction signals. To handle the unstructured and multi-modal nature of raw appeals, we employ a Vision–Language Model (VLM) pipeline to extract structured textual representations while preserving evidence traceability (Appendix~\ref{app:preprocessing}). Our evaluation spans over 260 production-level appeal cases, comprising a rigorous offline ablation study and a real-world online deployment. For the offline phase, we adopt an 80/20 chronological split of expert-adjudicated cases to construct the case knowledge base and assess reasoning fidelity; the training portion is used exclusively for retrieval indexing and graph construction, with no supervised fine-tuning performed on the underlying LLM at any stage.
Overturned cases exhibit significantly higher structural complexity than non-overturned ones (e.g., +144\% more verification actions on average); see Appendix~\ref{app:complexity} for a quantitative analysis.

\subsubsection{Evaluation Metrics} 




We assess system performance using \textbf{Decision Accuracy} and \textbf{Macro-F1} to measure alignment with human Checker judgments. Accuracy is defined as the exact match rate of system outputs $\hat{d}_q^{(C)} \in \{\approve, \reject, \rmi\}$ against ground truth. An \textsc{RMI} output is considered aligned if the system correctly identifies that essential verification actions (executed by Checkers via internal tools) cannot be grounded in the provided evidence, thereby correctly signaling a system boundary. Macro-F1 is reported to account for class imbalance, ensuring robust performance evaluation across minority outcomes like \textsc{Approve}. For online deployment, we monitor the \textbf{Cumulative Alignment Rate} and conduct qualitative failure analysis to categorize root causes of misalignment.

\subsubsection{Experiment Setting}
We implement the framework using \textbf{Claude Sonnet 4.0} as the uniform LLM backbone (via Amazon Bedrock) and \textbf{Amazon Titan Embed Text v2} (512-dim) for dense retrieval. During reasoning, the system retrieves the top-$K=20$ candidates via cosine similarity, subsequently refining them to $K'=5$ analogous precedents via an LLM-based relevance filter for factor alignment. To ensure reproducibility, we utilize greedy decoding (temperature $=0$) across all schema-guided agents. The Knowledge Base is populated exclusively from the training split, ensuring strictly inductive evaluation on the held-out test set. All baselines share identical backbone models and preprocessing to isolate the impact of the EAFD reasoning structure. No model parameters are updated during either offline evaluation or online deployment.

\subsubsection{Baselines}

\begin{table}[t]
\centering
\caption{Baseline taxonomy along four design dimensions.}
\label{tab:baseline_taxonomy}
\small
\begin{tabular}{lcccc}
\toprule
\textbf{Method} & \textbf{Uses} & \textbf{LLM} & \textbf{Action} & \textbf{RMI} \\
 & \textbf{History} & \textbf{Reasoning} & \textbf{Grounding} & \textbf{Enabled} \\
\midrule
LLM (Query-Only)    & \ding{55} & \ding{51} & \ding{55} & \ding{55} \\
LLM + RMI           & \ding{55} & \ding{51} & \ding{51} & \ding{51} \\
CBR (Majority Vote)  & \ding{51} & \ding{55} & \ding{55} & \ding{55} \\
LLM + CBR           & \ding{51} & \ding{51} & \ding{55} & \ding{55} \\
LLM + CBR + RMI     & \ding{51} & \ding{51} & \ding{51} & \ding{51} \\
\bottomrule
\end{tabular}
\end{table}

Table~\ref{tab:baseline_taxonomy} summarizes the baselines along four dimensions: whether the method leverages historical cases (\emph{Uses History}), employs LLM-based generation (\emph{LLM Reasoning}), requires verification actions to be grounded in evidence (\emph{Action Grounding}), and can output a \textsc{Request More Information} outcome (\emph{RMI Enabled}). We compare our framework against three primary paradigms: (1) \textbf{LLM (Query-Only)}: Direct decision prediction from case text without retrieval; (2) \textbf{CBR (Majority Vote)}: Outcome prediction based on the majority vote of top-$K$ retrieved cases; and (3) \textbf{LLM + CBR}: Standard RAG where retrieved cases serve as unstructured context for binary prediction. To isolate the impact of explicit action modeling, we evaluate \textbf{+ RMI} variants for both LLM and LLM+CBR settings. These variants enforce action grounding constraints, outputting \textsc{RMI} when verification steps cannot be grounded in evidence, contrasting with the binary classification of their standard counterparts.

\subsection{Main Results}

\subsubsection{Overall Comparison}

\begin{table}[h]
\centering
\caption{Overall adjudication performance under different reasoning
assumptions.}
\label{tab:overall_metrics}
\begin{tabular}{lccc}
\toprule
 & \multicolumn{3}{c}{\textbf{Overall Metrics}} \\
\cmidrule(lr){2-4}
Method & Acc. & Macro F1 & Macro Recall \\
\midrule
LLM (Query-Only)        & 0.708 & 0.604 & 0.728 \\
LLM + RMI     & 0.875 & 0.682 & 0.760 \\
CBR (Majority Vote)        & 0.370 & 0.250 & 0.322 \\
LLM + CBR     & 0.630 & 0.342 & 0.300 \\
LLM + CBR + RMI     & 0.667 & 0.354 & 0.342 \\
\midrule
Ours (EAFD)  & \textbf{0.958} & \textbf{0.867} & \textbf{0.889} \\
\bottomrule
\end{tabular}
\end{table}

Table~\ref{tab:overall_metrics} reports offline adjudication performance. \textbf{Ours (EAFD)} achieves superior performance (95.8\% Accuracy, 0.867 Macro-F1). In particular, naive retrieval (\textbf{LLM + CBR}) degrades performance compared to query-only reasoning (0.630 vs. 0.708), indicating that unstructured historical context introduces noise. In contrast, our structured action modeling effectively filters noise, achieving the strongest results by grounding reasoning in explicit verification paths.
Appendix~\ref{app:complexity} provides a concrete multi-stage case walkthrough illustrating how EAFD reasoning resolves Maker--Checker conflicts through action grounding.

\subsubsection{Ablation Studies}

\paragraph{(1) Effect of Action Grounding.}
We first examine the impact of explicitly modeling verification actions, by comparing LLM (Query-Only) with LLM + RMI, and LLM + CBR with LLM + CBR + RMI.
Across both query-only and retrieval-augmented settings, action grounding yields consistent improvements.
In the absence of retrieval, enabling action grounding improves accuracy from 70.8\% to 87.5\%.
When combined with retrieved cases, action grounding also improves performance from 63.0\% to 66.7\%, indicating that modeling verification actions remains beneficial even under retrieval.
These results suggest that many adjudication errors arise from missing or unverifiable verification steps rather than from language modeling alone.
This is further corroborated by the action hit rate analysis in Appendix~\ref{app:complexity}, where overturn cases achieve a higher hit rate (0.81 vs.\ 0.74), confirming that action grounding is especially critical for complex disputed cases.

\paragraph{(2) Effect of Structured Case Representation.}
We next examine whether retrieving historical cases as unstructured context is sufficient to leverage Maker--Checker correction signals, by comparing LLM (Query-Only) with LLM + CBR.
Naive case retrieval does not improve performance and in fact degrades accuracy (70.8\% vs.\ 63.0\%), highlighting the limitations of unstructured case reuse.
Historical appeal cases are heterogeneous and often contain case-specific artifacts or partially executed verification steps that do not transfer across instances.
In contrast, our full EAFD framework, which represents cases as structured graphs and integrates action grounding and factor-level alignment, substantially outperforms unstructured retrieval, achieving 95.8\% accuracy.
These results indicate that structured procedural representations are necessary for effectively reusing historical correction signals.

\paragraph{(3) Necessity of LLM-Based Reasoning.}
Finally, we assess whether adjudication can be solved through direct outcome reuse alone, by comparing the Pure CBR (Majority Vote) baseline with LLM-based variants.
The Pure CBR baseline performs poorly (37.0\% accuracy), indicating that outcome reuse without adaptive reasoning is insufficient.
Even among similar cases, differences in evidence availability and verification status require flexible interpretation.
These findings confirm that effective adjudication requires both structured case knowledge and LLM-based reasoning.

\subsection{Online Evaluation}

The framework has been deployed asynchronously in the production seller appeal workflow since January 17, 2026. It processes appeals in real-time, constructing EAFD graphs and retrieving precedents. The system acts as a safety guardrail: when verification actions cannot be grounded in evidence, it triggers \rmi rather than forcing a low-confidence decision.

\begin{figure}[t]
\centering
\begin{tikzpicture}
\begin{axis}[
    axis y line*=right,
    axis x line=none,
    ymin=0, ymax=150,
    ylabel={Cumulative Cases},
    ytick={0,50,100,150},
    tick label style={font=\small},
    label style={font=\small},
    width=\columnwidth, height=4.5cm, 
    xtick=\empty
]
\addplot[ybar, bar width=14pt, fill=orange!30, draw=none] coordinates {
    (1,5) (2,32) (3,68) (4,102) (5,136)
};
\end{axis}
\begin{axis}[
    xlabel={Date}, ylabel={Alignment Rate (\%)},
    ymin=95, ymax=100,
    xtick={1,2,3,4,5},
    xticklabels={Jan 17, Jan 20, Jan 25, Feb 1, Feb 5},
    tick label style={font=\small},
    label style={font=\small},
    width=\columnwidth, height=4.5cm, 
    grid=major, grid style={gray!20}
]
\addplot[thick, color=orange!80!black, mark=*, mark options={fill=orange!80!black}] coordinates {
    (1,100.0) (2,96.9) (3,96.7) (4,96.9) (5,97.1)
};
\end{axis}
\end{tikzpicture}
\caption{Cumulative alignment rate in production (Jan 17 -- Feb 5).}
\label{fig:online_alignment_time}
\end{figure}
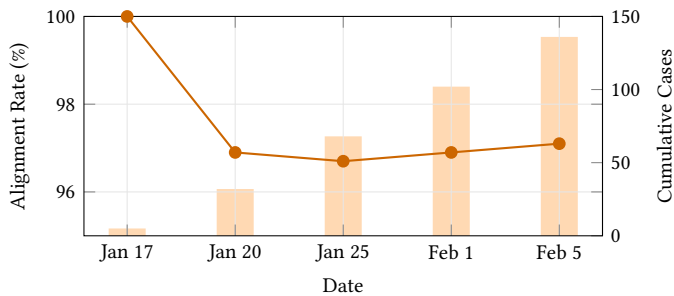

\begin{table}[h]
\centering
\caption{Online per-class performance over the first 136 cases.}
\label{tab:online_per_class}
\begin{tabular}{lcccc}
\toprule
Class & Precision & Recall & F1 & \# Case\\
\midrule
Approve (Reinstate) & 1.000 & 0.979 & 0.989 & 48 \\
Reject              & 0.988 & 0.966 & 0.977 & 87 \\
Request More Info   & 0.250 & 1.000 & 0.400 & 1 \\
\midrule
\multicolumn{1}{r}{Accuracy} 
 & \multicolumn{4}{l}{97.1\% (132 / 136)} \\
\bottomrule
\end{tabular}
\end{table}

\paragraph{Results \& Safety Analysis.}
Over four weeks (136 cases), the system achieved \textbf{97.1\% alignment} (Figure~\ref{fig:online_alignment_time}). Table~\ref{tab:online_per_class} highlights a critical safety characteristic: \textbf{100\% Precision on Approvals}. The system never reinstated a seller that the human Checker kept blocked. 

We analyzed the misaligned cases (including RMI outcomes) and found they were not reasoning failures, but correct identifications of the \textbf{System Boundary}. Specifically, human Checkers resolved these cases using internal tools unavailable to the LLM (e.g., \emph{Transparency Diagnostics Tool} queries, physical inventory bin checks). By outputting \textsc{RMI}, the system correctly signaled that the decision required information beyond the textual evidence context. This confirms that Action Modeling effectively prevents hallucination in high-stakes scenarios where information asymmetry exists.
Appendix~\ref{app:complexity} presents a representative multi-stage appeal (3 stages, 107 days) that illustrates how the system handles evolving evidence and cross-case dependencies in practice.

\section{Lessons Learned and Implications}
\label{sec:conc}

\textbf{Bridging the Multimodal Gap.}
A practical yet often overlooked challenge in industrial adjudication is the gap between raw case data and LLM-ready inputs. Real-world appeals comprise heterogeneous artifacts such as emails, invoices (PDFs), and chat screenshots, where LLMs struggle to ground textual claims in visual evidence. We found that direct retrieval over raw logs leads to severe hallucination, as the model cannot reliably associate text with its corresponding image evidence. To address this, we invested in a VLM-driven preprocessing pipeline (Appendix~\ref{app:preprocessing}) that converts visual evidence into structured textual representations with explicit source traceability. We first developed standardized case templates for human annotators, then applied VLMs to produce structured outputs from raw cases. This experience confirms that effective structured reasoning requires not just better models, but rigorous transformation of raw inputs into semantically linked representations.

\textbf{Policy-Aware Schema Adaptation.}
We observed that the topology of EAFD graphs is highly sensitive to Factor definitions: the same set of verification Actions can yield distinct graphs under different Factor interpretations. Rather than treating this as instability, we leveraged it as a design feature. By decoupling Factor definitions into a configurable module, the framework can dynamically adapt to shifting compliance standards (e.g., switching from lenient to strict risk controls) without retraining the underlying action extraction logic. This treats adjudication policy as a separable configuration layer, enabling flexible graph construction across diverse business lines and evolving operational requirements.

\textbf{From Fragmented SOPs to Data-Driven Standards.}
A primary driver of Maker-level inconsistency is the reliance on historically fragmented and often ambiguous Standard Operating Procedures (SOPs). Due to organizational evolution, multiple overlapping SOPs coexist, leading Makers to interpret abstract guidelines differently and produce inconsistent decisions. By aggregating granular Factors and Actions from successful Checker corrections, we aim to transition from top-down SOPs to data-driven standardization. Future work will focus on \emph{Factor Merging} to distill scattered reasoning paths into unified, canonical procedures. Unlike static rulebooks, these data-derived standards are grounded in concrete historical precedents, providing a self-evolving reference for human decision-makers.

\textbf{Graph Structure Validation.}
LLM-generated reasoning graphs are \emph{not} assumed to be structurally reliable by default. We introduce an explicit \emph{graph validation layer} that treats LLM outputs as \emph{proposals} rather than ground truth, enforcing EAFD schema invariants through deterministic checks on type compatibility, relational completeness, and cardinality constraints. This decouples probabilistic semantic extraction from deterministic structural enforcement. Details are provided in Appendix~\ref{app:validation}.

\section{Conclusion}
\label{sec:conclusion}

We presented a conflict-aware graph reasoning framework centered on the \textbf{EAFD schema} to address \emph{information asymmetry} in hierarchical review corrections. By modeling verification actions as inferential constraints, our framework transforms unconstrained text generation into structured reasoning over verifiable operations.

\textbf{Experimental Results.}
In offline evaluation, EAFD reasoning achieved a \textbf{95.8\%} alignment rate, significantly outperforming the 70.8\% LLM baseline, while online deployment maintained a \textbf{96.3\%} cumulative alignment rate over 136 production cases. The system achieved \textbf{100\% precision on approvals}, never reinstating a seller that human Checkers kept blocked. By identifying its system boundary through the \rmi capability, the framework correctly flagged missing evidence instead of hallucinating outcomes.

\textbf{Limitations \& Future Work.}
The system's efficacy depends on the richness of historical documentation, and factor definitions currently require domain-specific expert guidance. Future research will explore \emph{factor merging} to automate SOP derivation and apply the framework to other hierarchical domains like insurance and finance. By leveraging RMI signals for \emph{active learning}, we aim to further enhance system confidence and efficiency.

\bibliographystyle{ACM-Reference-Format}
\bibliography{references}

\clearpage
\appendix

\section{EAFD Schema Details}
\label{app:eafd_details}

This appendix provides detailed specifications of the EAFD schema components summarized in Section~\ref{sec:eafd_schema}.

\noindent\textbf{Node Type Definitions.}
The node set $\mathcal{V}_c$ comprises four disjoint subsets:
\begin{itemize}[leftmargin=*, nosep]
    \item $\mathcal{E}_c = \{ e_1, e_2, \ldots \}$ denotes \emph{evidence nodes}, each corresponding to an atomic factual snippet extracted from the case materials (e.g., seller statements, documents, or images). Evidence nodes are grounded in the original case text and retain source references for manual verification.
    \item $\mathcal{A}_c = \{ a_1, a_2, \ldots \}$ denotes \emph{action nodes}, representing concrete verification operations performed by reviewers, such as checking supplier authenticity or validating transaction records.
    \item $\mathcal{F}_c = \{ f_1, f_2, \ldots \}$ denotes \emph{factor nodes}, which encode abstract decision factors derived from action outcomes, such as compliance risks or credibility assessments. Factors reflect how verified information is interpreted in adjudication.
    \item $\mathcal{D}_c = \{ d^{(M)}, d^{(C)} \}$ denotes \emph{decision nodes}, corresponding to adjudication outcomes issued by the Maker and the Checker, respectively.
\end{itemize}



\noindent\textbf{Relation Type Specifications.}
Relations in $\mathcal{R}_c$ encode procedural dependencies between nodes:
\begin{itemize}[leftmargin=*, nosep]
    \item \textbf{Evidence--Action relations} $\mathcal{R}_{EA} \subseteq \mathcal{E}_c \times \mathcal{A}_c$ indicate which evidence items are used to support a verification action. This relation is many-to-many: a single action may rely on multiple evidence items, and the same evidence item may be reused across different actions.
    \item \textbf{Action--Factor relations} $\mathcal{R}_{AF} \subseteq \mathcal{A}_c \times \mathcal{F}_c$ capture how the outcome of an action contributes to a decision factor. Each action is constrained to be associated with exactly one factor. This one-to-one mapping ensures the semantic independence of factors, preventing a single verification action from ambiguously supporting multiple decision rationales.
    \item \textbf{Factor--Decision relations} $\mathcal{R}_{FD} \subseteq \mathcal{F}_c \times \mathcal{D}_c$ link decision factors to adjudication outcomes. These relations are annotated with a conflict indicator that specifies whether a factor supports or contradicts the associated decision. Each factor must be associated with exactly one decision.
    \item \textbf{Factor--Factor relations} $\mathcal{R}_{FF} \subseteq \mathcal{F}_c^{(M)} \times \mathcal{F}_c^{(C)}$ model how Checker-level factors evaluate, override, or revise factors previously considered by the Maker. This relation is constrained to be one-to-one: each Checker-level factor is required to correspond to exactly one Maker-level factor.
\end{itemize}


\section{VLM-Powered Data Preprocessing}
\label{app:preprocessing}

Raw case materials in industrial settings arrive as heterogeneous documents containing screenshots, pasted images, free-form annotations, and multi-turn conversation logs. Traditional OCR pipelines struggle with this diversity. The system addresses this challenge using Vision Language Models (VLMs).

\noindent\textbf{Single-Call Extraction.}
Rather than decomposing extraction into separate OCR, classification, and parsing stages, the pipeline provides all case images to the VLM simultaneously, instructing it to: (1) extract text while preserving semantic structure, (2) classify content by source (Maker vs.\ Checker artifacts), (3) identify key fields while filtering redundant details, and (4) annotate image sources for traceability. This unified approach achieves approximately 80\% token reduction compared to full OCR output while preserving decision-relevant information.

\noindent\textbf{Data Enhancement.}
The preprocessing stage applies targeted enhancements: \emph{image source annotations} maintain traceability from extracted text to original evidence; \emph{content type labels} distinguish primary evidence from interpretive annotations; \emph{URL explanations} provide context without requiring web fetches during graph construction.

\noindent\textbf{Quality Assessment.}
To prevent low-quality inputs from corrupting the knowledge base, an automatic quality assessment layer evaluates documents across six dimensions: completeness, logical consistency, evidence chain strength, decision explainability, documentation quality, and red flags. Cases falling below thresholds are flagged for manual review before entering the learning pipeline.

\section{Graph Structure Validation}
\label{app:validation}

LLM-generated reasoning graphs exhibit several failure modes that compromise structural integrity:

\noindent\textbf{Common Failure Modes.}
\begin{itemize}[leftmargin=*, nosep]
    \item \emph{Invalid edge directions}: Edges pointing from Decision to Factor instead of Factor to Decision.
    \item \emph{Missing mandatory relations}: Factors without any supporting Actions, or Actions without grounding Evidence.
    \item \emph{Cardinality violations}: Actions linked to multiple Factors (violating one-to-one constraint) or self-referential Factor-Factor edges.
\end{itemize}

\noindent\textbf{Validation Layer Design.}
The validation layer treats LLM outputs as proposals subject to schema-level constraints. The validation layer enforces EAFD invariants including: (1) type compatibility (edges only connect permitted node type pairs); (2) relational completeness (every Factor has at least one Action, every Action has at least one Evidence); (3) cardinality requirements (each Action maps to exactly one Factor).

\noindent\textbf{Repair Strategy.}
Graphs violating constraints are selectively repaired through targeted regeneration rather than end-to-end reprocessing. For instance, a missing Action-Factor edge triggers re-extraction of that specific relation while preserving validated portions of the graph.

This validation-and-repair mechanism ensures that reasoning graphs stored in the knowledge base satisfy strict structural guarantees, even when their semantic content is extracted by probabilistic language models.

\section{Appeal Complexity and Representative Case}
\label{app:complexity}

\subsection*{D.1 Quantitative Complexity of Overturn Cases}

We analyze a subset of 27 appeal cases from our production dataset as our testing data to
characterize the structural complexity of overturn decisions (\approve)
versus non-overturn decisions (\reject or \rmi). Among these 27 cases, 19 (70\%) resulted in overturns.

Overturn cases consistently exhibit higher complexity across all atomic
dimensions of the EAFD schema. On average, overturn cases involve
4.89$\pm$2.8 investigative actions (vs.\ 2.00$\pm$1.2 for non-overturn
cases, +144\%), 15.68$\pm$7.9 factual atoms (vs.\ 11.67$\pm$5.4, +34\%),
and 5.00$\pm$2.2 analytical findings (vs.\ 3.33$\pm$2.1, +50\%). The overall
atomic complexity, measured by total factor count, is 25.58$\pm$9.3 atoms
for overturn cases compared to 17.00$\pm$6.2 for non-overturn cases
(+50\%). These patterns quantitatively validate that successful appeals
require substantially deeper investigation and evidence accumulation to
overturn initial negative determinations.

Table~\ref{tab:action_analysis} summarizes action-level statistics,
highlighting how action grounding differs between overturn and
non-overturn cases.

\begin{table}[h]
\centering
\caption{Action-level analysis comparing overturn and non-overturn
appeal cases. ``Action Hit Rate'' measures the proportion of
system-recommended verification actions that semantically match at
least one Checker-documented action in the final review record.}
\label{tab:action_analysis}
\small
\begin{tabular}{lcc}
\toprule
 & \textbf{Overturn Cases} & \textbf{Non-Overturn Cases} \\
\midrule
Avg.\ Actions (System)          & 4.9  & 2.0  \\
Action Hit Rate                 & 0.81 & 0.74 \\
Cases with $\geq$1 Missing Action & 18\% & 41\% \\
\bottomrule
\end{tabular}
\end{table}

Overturn cases not only involve more actions, but those actions also align
more closely with human Checker behavior: the system's action hit rate is
higher and fewer overturn cases contain missing critical actions. This
suggests that successful reinstatements are driven by deeper,
better-aligned verification behavior rather than by superficial changes in
narrative alone.

\subsection*{D.2 Representative Multi-Stage Appeal Case Study}

\paragraph{Case Overview.}
This case illustrates the complexity of multi-stage appeal adjudication
under product quality enforcement. The appeal concerns an expired food
product violation (\texttt{PQ.EXPIRED\_PRODUCTS}) fulfilled by Amazon (FBA),
where the seller ultimately overturned the initial rejection after three
appeal stages over a 107-day period. The case involves evolving evidence,
explicitly conflicting facts, multiple verification actions, and
cross-case dependencies. While not the most extreme instance in the dataset,
this case represents a typical overturn scenario consistent with the
aggregate complexity statistics reported in Section~D.1.

\paragraph{Complexity Dimensions.}
This case exhibits several dimensions of adjudication complexity:
\begin{itemize}[leftmargin=1.2em]
    \item \textbf{Temporal depth:} Three appeal stages spanning 107 days.
    \item \textbf{Evidence evolution:} Atomic fact set expanding from 2 to 13
    with persistent contradictions.
    \item \textbf{Action depth:} Three independent verification actions
    (inventory review, warehouse verification, supplier contact).
    \item \textbf{Graph dependency:} Decision grounded in prior supplier
    verification from a separate appeal case.
    \item \textbf{Attribution ambiguity:} FBA fulfillment complicating
    responsibility assignment between seller and warehouse.
\end{itemize}

\paragraph{Stage 1: Initial Violation Detection (2025-Jan).}
The case was triggered by a customer complaint reporting receipt of a food
product expired by three months. Extracted atomic facts included 
\\ \texttt{EXPIRED\_PRODUCT\_RECEIVED} and \texttt{FBA\_FULFILLMENT}. No
verification actions beyond complaint validation were performed. Following
zero-tolerance food safety policy, the initial decision was
\textsc{Reject}.

\paragraph{Stage 2: First Appeal (2025-Jan).}
The seller submitted purchase orders, inventory records, and FIFO
documentation from the supplier. This introduced new supporting facts
(\texttt{PURCHASE\_ORDERS\allowbreak\_PROVIDED},
\texttt{FIFO\_DOCUMENTATION\allowbreak\_PROVIDED}) and triggered an\\
\texttt{INVENTORY\allowbreak\_REVIEW\allowbreak\_CONDUCTED} action. The resulting finding
\texttt{FIFO\allowbreak\_MANAGEMENT\allowbreak\_VALIDATED} mitigated concerns of systematic
inventory negligence, leading to a provisional overturn.

\paragraph{Stage 3: Second Appeal (2025-Feb).}
Additional third-party warehouse inspection evidence was provided. A
\texttt{WAREHOUSE\allowbreak\_VALIDATION} action confirmed proper storage
conditions, upgrading the evidentiary strength from seller-provided
documentation to independent validation. The appeal again passed review.

\paragraph{Challenge Review and Cross-Case Validation (2025-May).}
The final review consolidated all evidence and performed direct supplier
verification. The complete fact set comprised 13 atomic facts, including one
contradicting fact (\texttt{EXPIRED\_PRODUCT\_RECEIVED}) and multiple
supporting facts such as \texttt{TRUSTED\_SUPPLIER\_STATUS} and
\texttt{HIGH\_SALES\_VOLUME\_WITHIN\_SHIP\_DATE}. Statistical analysis
(1 defect out of 4,523 compliant transactions; 0.022\%) supported an
\texttt{ISOLATED\_INCIDENT} finding. The system additionally retrieved nine
high-similarity precedent cases, all resolving to \textsc{Approve}.

\paragraph{Conflict Resolution via EAFD Reasoning.}
The conflict between objective violation evidence and systemic compliance
signals was resolved through explicit action grounding rather than direct
inference. Verification actions produced concrete findings, quantitative
risk assessment contextualized defect severity, and cross-case supplier
validation eliminated single-case bias. The final decision was
\textsc{Approve}.

\end{document}